\documentclass{article}

\PassOptionsToPackage{numbers, compress}{natbib}
\usepackage[preprint]{neurips_2026}

\usepackage[utf8]{inputenc} % allow utf-8 input
\usepackage[T1]{fontenc}    % use 8-bit T1 fonts
\usepackage{hyperref}       % hyperlinks
\usepackage{url}            % simple URL typesetting
\usepackage{booktabs}       % professional-quality tables
\usepackage{amsfonts}       % blackboard math symbols
\usepackage{nicefrac}       % compact symbols for 1/2, etc.
\usepackage{microtype}      % microtypography
\usepackage{xcolor}         % colors

\usepackage{graphicx}
\usepackage{booktabs}
\usepackage{multirow}
\usepackage{wrapfig}
\usepackage{amsmath}
\usepackage{wrapfig}
\usepackage{titlesec}
\usepackage{pifont}
\title{An LMM for Precisely Grounding Elements in Documents}

\author{Yijian Lu$^{1}$ \hspace{0.5em} Chuangxin Zhao$^{2}$ \hspace{0.5em} Kai Sun$^{1}$ \hspace{0.5em}Lei Hou$^{1}$ \hspace{0.5em} Ji Qi$^{1*}$ \hspace{0.5em} Juanzi Li$^{1*}$\\
  $^{1}$Department of Computer Science and Technology, Tsinghua University\\
  $^{2}$Institute of Automation, Chinese Academy of Sciences \\
\texttt{\{luyj24, qiji\}@mails.tsinghua.edu.cn}
}

\begin{document}

\maketitle
\renewcommand{\thefootnote}{}  % 临时清空编号显示
\footnotetext{*Corresponding Authors.}
\begin{abstract}
%Visual grounding enhances the interpretability and reasoning capabilities of Large Multimodal Models (LMMs) in Visual Question Answering (VQA) by localizing relevant visual regions.
Visual grounding in documents is a crucial ability for Large Multimodal Models (LMMs) in areas such as document understanding, deep research and document error detection. However, existing approaches exhibit poor grounding precision in text-rich document images, often failing to accurately locate the critical document elements needed for reliable reasoning. To address this gap, we introduce PreciseDoc, an LMM specifically designed for precise element grounding and can be further optimized for Document VQA tasks. Specifically, to enhance the basic localization capability, we construct challenging training data by two pipelines capable of mass-producing high-quality documents with paired metadata of fine-grained coordinates, including synthetic hand-filled documents with camera effects. The model develops more real-world functions beyond straightforward localization of single text, such as locating personal information from CVs. Furthermore, we introduce a training paradigm for visual grounded reasoning where the grounding and reasoning are supervised jointly with reinforcement learning to improve the contribution of the grounded evidence. A comprehensive evaluation on various benchmarks demonstrates the advantage of the proposed data and methods in document spatial grounding and document understanding.
\end{abstract}

\section{Introduction}
\label{sec:intro}

Visual grounding is the ability of LMMs to refer to relevant visual regions based on natural language instructions, which reflects the level of image understanding. Moreover, it has been extended to VQA tasks as part of the reasoning process. Specifically, when the textual reasoning of an LMM commits to a specific region in an image, it must include the corresponding spatial coordinates \cite{pantazopoulos2025towards}. By emphasizing such verifiable grounding information, this capability, referred to as visual grounded reasoning, can prevent over-reliance on textual descriptions alone. Existing research has demonstrated its effectiveness in natural-image VQA settings \cite{wang2025traceable}, where interpretability and reasoning capabilities are significantly enhanced by accurate visual evidence localization. 

However, document images present a significantly more challenging scenario than natural images in this setting and remain a largely underexplored area in the literature. First, existing LMMs often fail to perform precise visual grounding in documents, a task that requires locating a given element (words, phrases, etc.). This is because document images are characterized by high element density and variety, rich template diversity and layout complexity. Such deficient grounding capability, if directly incorporated into visual grounded reasoning where the critical evidence the model needs to locate is often confined to a small, localized region, can potentially cause performance degradation. In that case, the LMM may focus on a unrelated region and produce a false final answer.
%In document understanding, the critical evidence is often confined to a small, localized region within the visual content. Consequently, existing models often fail to accurately localize specific elements within document images. Directly incorporating this deficient grounding capability into visual grounded reasoning can backfire, potentially harming answer correctness as a result of localization errors. 
Therefore, a two-step approach that first enhances foundational grounding ability before integrating it into reasoning is in urgent need.

To bridge this gap, we introduce \textbf{PreciseDoc}, an LMM specifically designed to accurately identify and localize key elements within document images, and can be further optimized for the complex task of visual grounded reasoning.
%effectively integrate grounded information into subsequent reasoning steps. 
To first solve the basic document grounding bottleneck, we construct two document generation pipelines that support mass production of high quality documents. In addition to content-rich resume PDFs, the pipelines produce synthetic hand-filled documents with camera effects. Moreover, accurate location metadata are generated simultaneously with the document that facilitates subsequent data curation for visual grounding and tasks. Compared to existing datasets such as Doclocal4K \cite{hu2024mplug}, the curated data present a steeper visual challenge by requiring multiple locations in a single question.
%two information extraction datasets: one based on multi-page native PDF files requiring the coordinates of elements belonging to the same type; the other based on synthetic handwritten documents with camera effects requiring coordinates of elements by content category. Both datasets pose more challenges than pure grounding on single-page PDF files and better align with real-world needs.
After training, PreciseDoc develops more real-world functions than straightforward text grounding, such as locating critical information from documents.

Furthermore, we introduce the first complete training pipeline with reinforcement learning that equips LMMs the ability of visual grounded reasoning in documents. In the cold-start stage, we curated a 27K VQA dataset on various document types such as academic papers and posters, where the thinking process includes the coordinates of key regions.
%of both element-level (images, paragraphs) and token-level (words, phrases). 
%The former level refers to complete, high-level document elements, such as images, paragraphs, tables, and charts, which are typically the primary units of information in a dense academic paper. The latter denotes much smaller, granular units, such as individual words or phrases, which are particularly critical for understanding the concise, visually-driven content of a poster or a standalone chart. 
Then, the reasoning model, referred to as \textbf{PreciseDoc-Reasoner}, underwent joint reinforcement learning training that supervises both reasoning accuracy and intermediate grounding through an answer reward and an IoU-based reward. In the grounding reward, unlike previous design using a many-to-many mapping, which is highly susceptible to metric inflation, we employ an optimal one-to-one assignment between predictions and ground truths via the Hungarian Algorithm \cite{https://doi.org/10.1002/nav.3800020109}. To avoid redundant predictions, we introduce a strict length penalty to explicitly penalize any unassigned generated boxes. In our experiments with pure localization benchmarks, the proposed PreciseDoc demonstrated improved 
grounding ability. For visual grounded reasoning, the PreciseDoc-Reasoner is capable of producing more precise intermediate grounding locations, and answer accuracy comparable to other modern LMMs. 

In summary, our contributions are threefold. (1) We introduce two document generation engines capable of producing high-quality documents and more challenging grounding data. (2) We introduce PreciseDoc, an LMM with SOTA performance in the task of document text grounding and real-world grounding functions other than direct text grounding. (3) We further optimize and obtain the PreciseDoc-Reasoner for visual grounded reasoning in documents, with the first practice of a complete training pipeline from SFT to RL. During RL, the grounding performance is jointly supervised with reasoning through an IoU reward with the Hungarian Algorithm and a length penalty.

\section{Related Work}

\subsection{Document Understanding in LMMs}
The field of document understanding has shifted from traditional supervised methods to generative approaches driven by recent advances of LMMs without reliance on OCR tools. Typically, mPLUG-DocOwl \cite{ye2023mplug} emerged as one of the specialized models achieving strong performance for text-dense scenarios, as well as UniDoc \cite{feng2023unidoc}, UReader \cite{ye2023ureader}. More recently, several LMMs including TextMonkey \cite{liu2026textmonkey}, InternLM-XComposer4KHD \cite{dong2024internlm}, and InternVL 1.5 \cite{chen2024far} have employed techniques like shape-adaptive cropping, shifted windows, and increased image tiles to better process high-resolution document images. However, a critical limitation persists that the produced responses cannot be accurately and visibly linked to specific regions within the document. This disconnection between understanding and localization hinders fine-grained readability and interpretability.

\subsection{Visual Grounded Reasoning in LMMs} %natural image; document image; reasoning with visual grounding
The concept of "thinking with images" has already motivated numerous efforts to dynamically incorporate visual grounding into reasoning LMMs, beyond merely an independent ability of Referring Expression Comprehension (REC) \cite{qiao2020referring,chen2025revisiting,zhang2019referring,kazemzadeh2014referitgame, yu2016modeling}, or a decoupled preprocessing step before the actual answer generation \cite{qi2024cogcom,mondal2024kam,liu2024chain, shao2024visual}. Early approaches such as VRG \cite{wang2025vgr}, construct SFT data to enable the output of relevant grounding coordinates. Later approaches integrate vanilla reinforcement learning \cite{sutton1998reinforcement} to encourage models to generate spatially-grounded outputs \cite{park2025dip,fan2025grit, liu2025visionreasoner}. Under this framework, various interaction mechanisms have been explored. For instance, DeepEyes \cite{zheng2025deepeyes} invokes an image zoom-in function during reasoning based on generated coordinates, while Ground-R1 \cite{cao2025ground} facilitates multiple turns of grounding and reasoning. While the reward design of these works focus only on the final answer, TreeVGR \cite{wang2025traceable} proposes to supervise the intermediate bounding boxes through an additional IoU-based grounding reward, and achieves more precise localization.

%openai-o3 (image manipulation during reasoning); VGR (construct sft data); DIP-R1 (RL for visual detection); visual-cot (before reasoning); grit (vanilla RL， where format reward encourages model to output bounding boxes); deepeyes (invoke an image zoom-in function during reasoning); Visionreasoner (multi-task reinforcement learning); Ground-R1 (rl, multiple turns of grounding and reasoning); TreeVGR (RL additionally supervising generated bounding boxes)

% DOGR (fine-tuned model based on a sft dataset, bounding box coordinates of four text granularities, (word, phrase, line, paragraph))，TRIG (sft,embedding-based method: we fine-tune an MLLM as an encoder and estimate the grounding area using image-text embedding similarities.)，BBox-docVQA (the first large-scale bounding-box–grounded DocVQA dataset, box types include text, image and table)，boundingDocs (combining several public DocVQA datasets and directly associate answers with OCR-extracted tokens)
Visual grounded reasoning in documents, however, is a largely underexplored area with limited methods. Specifically, \citet{giovannini2025boundingdocs} fine-tuned several LMMs on a reformed datasets combining public VQA data where answers are directly associated with their exact positions. Similarly, DOGR \cite{zhou2025dogr} conducts model fine-tuning based on a novel VQA dataset of three document types, containing token-level bounding box in each answer. Inspired by ColPali \cite{faysse2024colpali}, TRIG \cite{li2025towards} proposes an embedding-based method that trains a visual encoder to predict the grounding area based on image-text embedding similarities. However, no complete training pipeline has been designed and implemented involving both supervised fine-tuning and reinforcement learning.

%\section{Method}
%In this section, we introduce our PreciseDoc. Specifically, the model establish a foundational document grounding capability by leveraging several localization datasets on both PDF files and synthetic handwritten documents containing elements of varying granularities and types (Section \ref{method_grounding}). The subsequent Section \ref{method_reasoning} contains the cold-start initialization and reinforcement learning designed to progressively unlock and strengthen the visual grounded reasoning capability, $i.e.$, generating precise bounding box coordinates of relevant visual evidence during reasoning.

\section{Visual Grounding}
\label{method_grounding}

\begin{figure}[t]  % h表示放在此处
    \centering     % 图片居中
    \includegraphics[width=0.99\textwidth]{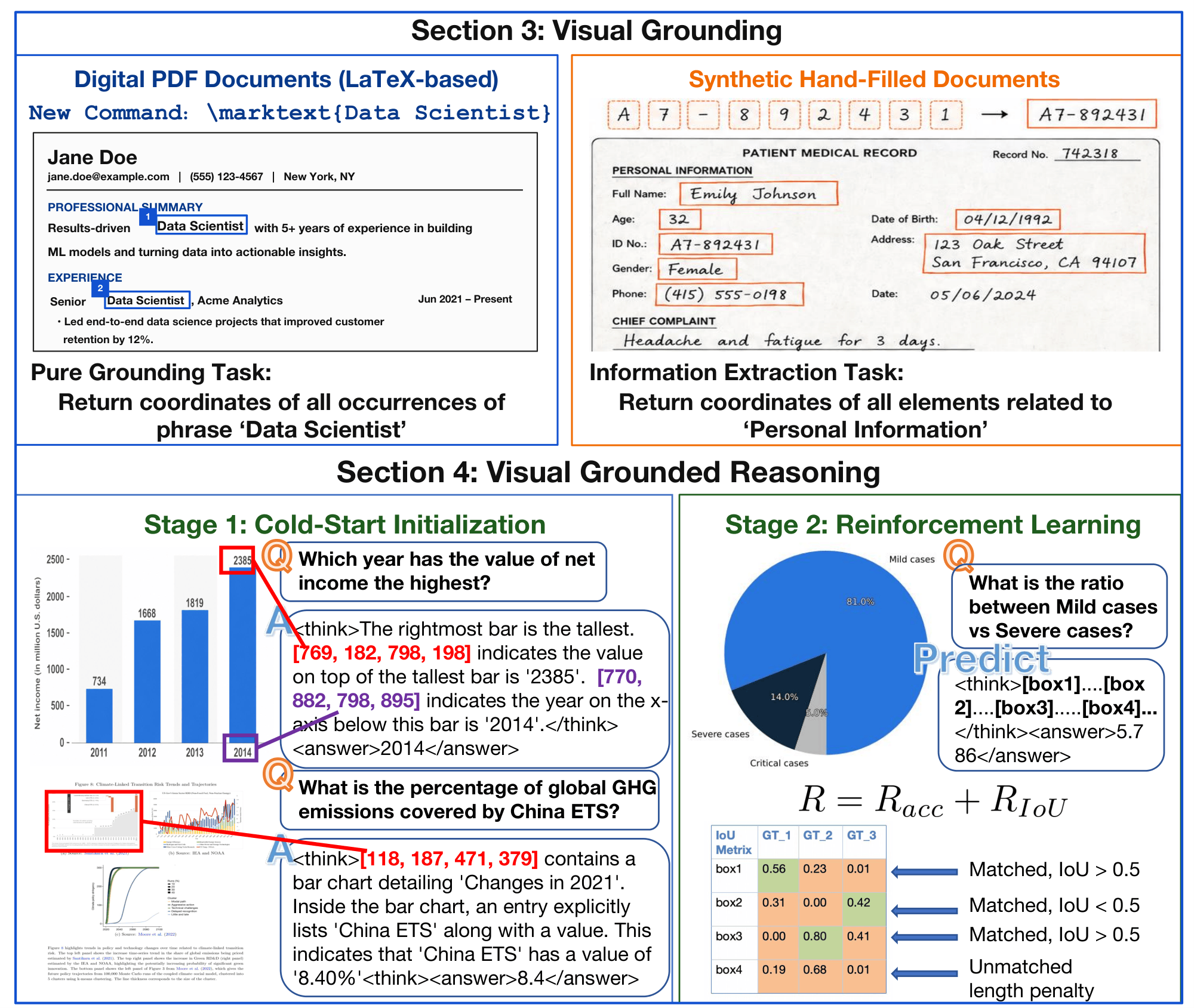}
    \caption{Illustration of Section \ref{method_grounding} and Section \ref{method_reasoning}. The upper part of the figure presents the two document generation engines and two types of grounding tasks. The spatial metadata of Latex-based PDFs are obtained by the designed command of \texttt{\textbackslash marktext} while the handwriting for hand-filled documents is obtained through single-character assembly. The bottom part presents the training pipeline of PreciseDoc-Reasoner. In stage 1, the model is instructed to generate token-level (the first question in stage 1) and element-level (the second question in stage 1) bounding boxes during reasoning. In stage 2, when calculating the grounding reward, we will first obtain the optimal one-to-one matching via Hungarian Algorithm, then compute the rate of matched pairs whose IoU exceeds a threshold. Finally, the reward is reduced by a length penalty from any unmatched boxes. }  % 图注
    \label{fig:example}         % 标签，用于引用
\end{figure}

In this section, we introduce how the proposed PreciseDoc establishes a foundational document grounding capability before the visual grounded reasoning in Section \ref{method_reasoning}. Generally, we construct training data by two document generation engines: one compiles LaTeX source codes, and the other produces synthetic hand-filled documents with camera effects. Both pipelines jointly produce paired documents and metadata, facilitating the direct acquisition of training data for grounding.

\subsection{Latex-based Data Engine}
The construction of a training dataset on native PDFs suffers from the fact that good-quality PDFs can only be collected, but cannot be mass-produced since even the most advanced VLMs produce repeated and formulaic templates. Furthermore, the accuracy of coordinates obtained by document parsing software is insufficient. This led us to consider using LaTeX source code to generate PDFs. This is because LaTeX templates come in a wide variety and support mass production. Most importantly, accurate coordinates can be obtained by inserting commands during the compilation.

\subsubsection{Template Collection and Manipulation}
We narrow our focus to the résumé templates, which feature rich information and complex structure. First, we manually obtain a collection of open-source Latex source codes and filter those cannot be successfully compiled on local environment, and those resulting in multiple pages. For mass production, we input the original template file to the VLM and demand it to produce a number of complete files with contents replaced and layouts untouched.

Then, we design a new command \texttt{\textbackslash marktext} in Latex that stores the coordinates of wrapped texts during compilation. Specifically, it measures the width, height, and depth of the wrapped text, utilizes the inborn \texttt{\textbackslash zsavepos} to obtain the bottom‑left page coordinate and writes these values to the .aux file. The coordinates are in points (pt) so conversion to pixels is performed following the file generation. We filter out bad cases due to compilation errors. For an accuracy check of the remaining files, we crop the image using the coordinates and deploy an OCR model to verify the consistency between the wrapped text and the cropped image.

\subsubsection{Dataset Construction by task}
\textbf{Pure Grounding.}
For each image, we first obtain the top-k most frequent n-grams with length more than $N$ in the document. Then we wrap all occurrences of each n-gram with the command \texttt{\textbackslash marktext} in the Latex code, and compile to obtain their coordinates in the metadata file. Each training data will require the localization of all occurrences of a given phrase.

\textbf{Information Extraction.}
For each document, we query the VLM with the complete Latex code and obtain the main categories that each group a number of elements. For example, category "Work Experience" contains related elements of work title, duration, responsibilities, etc. Then, we randomly select one category as the sample question and store the coordinates of elements in that group.

\subsection{Handwritten Data Engine}
Due to variations in handwriting styles, as well as quality degradation caused by camera scanning, handwritten documents present a more challenging visual grounding scenario compared to native PDFs. Moreover, to the best of our knowledge, there is no existing dataset for this setting. Therefore, inspired by PaddleOCR \cite{cui2026paddleocrvl15multitask09bvlm}, we decide to synthesize our own handwritten document dataset.

\subsubsection{Handwritten Document Generation}
We develop a fully automated pipeline to generate handwritten documents paired with detailed element metadata due to the lack of existing datasets. First, we query a VLM to generate pairs of content list and layout that together are rendered as an empty document with only field names and space reserved for handwritten values. To improve the layout diversity, the offset of the document content's starting point from the page top is randomly initialized. Then, we build a handwriting generator that assembles individual handwritten character samples together and outputs handwritten images of a given phrase or sentence. Specifically, the NIST \cite{Grother1995NISTSD} datasets provide complete sets of handwritten characters and digits from different individuals. As long as the same handwriting style is used consistently when assembling characters within the same document, a high level of fidelity can be attained. After that, we query the VLM to generate values for each field name while ensuring that all characters produced exist in the handwriting database. The handwritten images are inserted into the reserved spaces. However, due to variations in the length of the values, the pre-reserved spaces do not accurately reflect the actual bounding boxes of the handwritten content. Therefore, we obtain tight bounding boxes based on the actual sizes of the inserted images. At last, to better simulate the way handwritten documents are captured and stored, we apply various post-processing effects to the documents, such as filters that cause image quality degradation.

\subsubsection{Dataset Construction by task}
\textbf{Pure Grounding.}
For each image, we randomly select one handwritten content from the metadata file and store its corresponding bounding box coordinate as training output.

\textbf{Information Extraction.}
For each image, we randomly select one category, e.g. Personal Information, and store the coordinates of all handwritten elements whose field name falls into that category.

\section{Visual Grounded Reasoning in Documents}
\label{method_reasoning}
This section contains the process of further optimizing PreciseDoc from Section \ref{method_grounding} to \textbf{PreciseDoc-Reasoner}. The optimization begins with the cold-start initialization and reinforcement learning designed to progressively unlock and strengthen the visual grounded reasoning capability, $i.e.$, generating precise bounding box coordinates of relevant visual evidence during reasoning.

\subsection{Cold-Start Initialization}
\label{method_cold_start}
A cold-start supervised fine-tuning stage before the end-to-end reinforcement learning has been proven effective by \cite{wang2025traceable} to reduce computational intensity and experimentation inaccessibility. Specifically, the cold-start stage aims to equip the model with the visual grounding ability during the reasoning process, i.e., generating bounding box coordinates upon mentioning relevant document regions. Therefore, the training sample for cold start should include single or multiple image inputs, a question, reasoning trajectories with corresponding bounding boxes and a final answer, to guarantee the possession of grounding capacity prior to RL training.

\subsubsection{Dataset Construction}

We base our SFT dataset on part of the BBox-DocVQA training set \cite{yu2025bbox} and DOGR training set \cite{zhou2025dogr}, which provide final answer and ground-truth bounding boxes. The bboxes in BBox-DocVQA primarily include larger document elements such as paragraphs, images and tables. On the other hand, the bboxes in TRIG are token-level or word-level, containing mostly words, phrases and sentences. The goal here is to generate valid and detailed intermediate reasoning steps involving ground-truth bounding boxes.

\textbf{Step 1 BBox Preprocessing.}
The preprocessing for bboxes consists of two parts: bbox expansion and bbox caption generation. The dataset size of DOGR is significantly larger than BBox-DocVQA. To address the unbalance, bbox expansion is applied to convert token-level bboxes into element-level ones, thereby augmenting the data volume. We apply MinerU \cite{wang2024mineru} to the document images in the DOGE dataset to obtain element-level OCR objects. When a token-level ground truth bounding box was either spatially encompassed by a parsed bounding box, or when the text within the ground truth bounding box constituted a substring of the text within the parsed bounding box, we replaced the original bounding box with the corresponding MinerU-parsed bounding box for that sample. Often, multiple ground truth bounding boxes end up belonging to the same element-level bounding box.

To facilitate better understanding and intergration of bboxes into reasoning step generation from an instructed VLM, here we annotate the actual content within the ground-truth bboxes according to the element types. Specifically, we directly append the OCR texts if the bbox is token-level or the content is text, such as paragraphs and table. For images, we generate a detailed description, and the complete content in case of structured images such as charts.

\textbf{Step 2 Thinking Generation.}
We utilize a more advanced VLM, GLM-4.6V to obtain the thinking process. Specifically, we input the images and question, along with the ground-truth bboxes annotated with content information to the VLM, and instruct the VLM to generate the thinking content and a final answer. During the VQA process, the VLM is required to examine the given bboxes and make use of them as much as possible to derive the answer. It should be noted, however, that not all bounding boxes are relevant, owing to quality concerns in a small subset. The bounding boxes serve as substantive subjects that actively participate in the reasoning (e.g., "Bbox 1 states clearly that"), rather than as supplementary annotations(e.g., "The number of deals is 4 [Bbox 1]"). Finally, to ensure the accuracy of the reasoning traces, we compare the final answers generated by the VLM against the ground truth answers. Only those samples where the two answers are consistent will be retained in the final SFT dataset.

It is worth noting that all grounding bounding boxes within a single response must be exclusively at either the element level or the token level. To differentiate the granularity of the grounding boxes output in the reasoning trace, we specifically state this requirement in the question prompt and elaborate on their distinct characteristics, such as differences in content within the bbox and relative size.

\subsection{Reinforcement Learning}
\label{method_rl}
We proceed to reinforcement learning to refine the model's reasoning trajectories by supervising its intermediate grounding outputs. Specifically, the generated bounding box coordinates are inherently structured and can be quantitatively evaluated through geometric metrics, $i.e.$, the Intersection-over-Union (IoU) with the ground-truth annotations. This reward component ensures explicit supervision over the visual evidence produced by the model.

\subsubsection{Reward Design}
The total reward $R$ consists of two parts: an answer reward $R_{acc} \in \{0,1\}$ and a dual Intersection-over-Union (IoU) reward $R_{IoU} \in [0,1]$:

\setlength{\abovedisplayskip}{1pt}
\setlength{\belowdisplayskip}{1pt}
\begin{equation}
R = \omega R_{acc} + (1-\omega)  R_{IoU},
\end{equation}

where $\omega$ is the weight coefficient that controls the importance of the two rewards. There is no format reward regulating the model to generate reasoning and final answer in special tokens. Instead, it is implicitly integrated into the computation of the $R_{acc}$ and $R_{IoU}$. Specifically, only the predicted answer wrapped in special tokens, and bounding boxes in correct JSON format during thinking phrase will proceed to subsequent reward calculation. For answer accuracy, we utilize an online reward model to judge whether the prediction is consistent with the ground-truth answer given the question.

The dual IoU reward $R_{IoU}$ measures the quality of the predicted bounding boxes with ground-truths. Specifically, $R_{IoU}$ is an average of the grounding score $R^k_{IoU}$ between the predicted boxes $\lbrace \hat{b}^{k}_i \rbrace_{i=1}^{N}$ and ground-truth boxes $\lbrace b^{k}_j \rbrace_{j=1}^{M}$ in each page $k$ since the given document is multi-page. To calculate the grounding score inside page $k$, we first compute a pairwise IoU Matrix $\mathbf{M}_{(i,j)} = \mathbf{IoU}[\hat{b}^{k}_i,b^{k}_j]$ between prediction and ground-truths. Then we obtain the optimal one-to-one assignment via the Hungarian Algorithm \cite{https://doi.org/10.1002/nav.3800020109}, which maximizes the total IoU across matched pairs while ensuring that each prediction maps to at most one ground-truth, and each ground-truth maps to at most one prediction. Finally, the initial $R^k_{IoU}$, computed as the fraction of assigned pairs whose IoU exceeds a predefined threshold $\epsilon$, will be given a length penalty $- \frac{\left| |\hat{b}^{k}_i| - |b^{k}_j| \right|}{|\hat{b}^{k}_i|}$ to handle the unmatched boxes. 

This is a stricter and more faithful evaluation of localization quality than the grounding score calculation implemented with a many-to-many mapping \cite{wang2025traceable}, which takes the average of the maximum IoU between all predictions and each ground-truth, i.e., $R^k_{IoU} = \frac{1}{M}\sum_{j=1}^{M}max_iIoU(\hat{b}^{k}_i, b^{k}_j)$. For example, a prediction contains 5 nearly identical boxes closely matching a single ground-truth will result in an inflated precision, while under our reward design, only one will be matched, leaving the rest 4 boxes as penalty.

\subsubsection{Data Construction}
%As discussed above, during reinforcement learning, the grounding score needs to be calculated as part of the reward. Therefore, the training data must include ground-truth bounding box. 
After cold-start initialization, the SFT checkpoint has been equipped with fairly strong document reasoning capability, and simple questions will not benefit the training any more. To this end, we filtered out the questions that the cold-start model could already answer with 100\% accuracy by rolling out its responses on the original Bbox-DocVQA and DOGR training set, retaining 26K challenging samples as the final RL dataset.

\section{Experiments}

\subsection{Experiment Setup}

\subsubsection{Benchmarks and Metrics}
For systematic evaluation, we benchmark our models on both visual document grounding and document understanding. Specifically, DocLocal4K \cite{hu2024mplug} is a multi-granular text localization dataset evaluating text grounding abilities with targeted text in the input prompt. Furthermore, TRIG \cite{li2025towards} and BBox-DocVQA \cite{yu2025bbox} are two newly-proposed datasets benchmarking not only answer generation, but also evidence localization from intermediate visual grounding with token-level and element-level ground-truth bbox coordinates respectively. 

On the DocLocal4K dataset, grounding accuracy is assessed via IoU@0.5. 
TRIG assesses evidence grounding by calculating the IoU between the union of the predicted boxes and the union of the ground truth boxes. Meanwhile, BBox-DocVQA utilizes LLM-as-judge to rate semantic correctness of reasoning outputs in addition to IOU calculation of the spatial coordinates of three types: Single-Page Single-BBox; Single-Page Multiple-BBox; Multi-Page Multiple-BBox. 

\subsubsection{Implementation Details}
The whole training process for PreciseDoc and PreciseDoc-Reasoner is provided in Appendix \ref{appendix:train}.All training and evaluation experiments are run on NVIDIA H100 GPUs. Answer accuracy during reinforcement learning and evaluation on BBox-DocVQA is judged using Qwen-2.5VL-72B-Instruct and DeepSeekv3.1 respectively. The weight coefficient $\omega$ during RL is set to 0.8 to place a greater emphasis on the accuracy of final answers. IoU threshold $\epsilon$ is set to 0.5.

\subsection{Experiment Results}

We evaluate multiple state-of-the-art private models, including Gemini-3-Flash, Gemini-3-Pro from Google DeepMind \cite{team2023gemini}, GPT-4o, GPT-5 \cite{singh2025openai}, GPT-5.2 from OpenAI, as well as Doubao-Seed-2-0-Pro \cite{guo2025seed1}. The open-source general models, we include LLaVA \cite{liu2023visual}, InternVL3 series \cite{zhu2025internvl3}, Qwen3 and Qwen3.5 \cite{qwen3.5} series of various parameter scales. Furthermore, we incorporate specialized models that excel particularly in text grounding, such as DocOwl-1.5 and Monkey-Chat \cite{li2024monkey}.
% \usepackage{booktabs}
% \usepackage{multirow}
\begin{table}[]
\centering
\scriptsize
\caption{Performance comparison on DocLocal4K across various private and open-source models.}
\label{doclocal4k_table}
\begin{tabular}{@{}l|ccccc@{}}
\toprule
\multirow{2}{*}{Model} & \multicolumn{5}{c}{Text Localization} \\ \cmidrule(l){2-6}
                       & word & phrase & line & paragraph & ALL \\ \midrule
Gemini-1.5-Pro         & 6.55 & 3.69 & 4.37 & 5.98 & 5.18 \\
Gemini-2.0-Flash       & 4.03 & 5.54 & 7.36 & 9.69 & 6.49 \\
Gemini-2.5-Flash       & 8.57 & 8.86 & 10.74 & 10.72 & 9.65 \\
Gemini-2.5-Pro         & 23.19 & 39.11 & 38.57 & 10.10 & 27.91 \\
Gemini-3-Flash         &27.06 & 34.87&36.38 & 30.10&31.95 \\
GPT-4o                 & 6.89 & 2.77 & 3.18 & 8.25 & 5.27 \\
GPT-5.2                 & 1.01 & 0.37 & 2.39 & 14.64 & 4.28 \\
InternVL2.5-8B         & 8.89 & 5.19 & 2.60 & 12.61 & 7.21 \\
InternVL3-8B         & 1.85 & 2.95 & 2.98 & 1.03 & 2.21 \\
Qwen2.5-VL-7B          & 20.84 & 13.84 & 25.25 & 27.01 & 21.51 \\
Qwen3-VL-8b-thinking   & 18.66 & 8.12 & 18.29 & 23.09 & 16.89 \\
Qwen3.5-9B   & 15.46 & 5.90 & 17.50 & 19.38 & 14.40 \\
Qwen3.5-397B   & 36.64 & 36.72 & 46.12 & 79.38 & 48.66 \\
Doubao-Seed-2.0-Pro  & 45.88 & \underline{43.91} & 53.48 & 77.73 & 54.45 \\
DocOwl-1.5             & \textbf{70.42} & \textbf{76.38} & \underline{85.88} & \underline{91.34} & \underline{80.38} \\ \midrule
\textbf{PreciseDoc}                       & \underline{67.73} & \textbf{76.38} & \textbf{89.46} & \textbf{97.11} & \textbf{81.79} \\ \bottomrule
\end{tabular}
\end{table}

% \usepackage{booktabs}
% \usepackage{multirow}
\begin{table}[]
\centering
\scriptsize
\caption{Performance comparison of grounding and answer accuracy on TRIG. Evaluation setting requires models to generate bounding box coordinates from scratch and reports IoU.}
\label{trig_table}
\begin{tabular}{@{}l|ccccc@{}}
\toprule
\multicolumn{1}{c|}{\multirow{2}{*}{Model}} & \multicolumn{5}{c}{Grounding Accuracy} \\ \cmidrule(l){2-6} 
\multicolumn{1}{c|}{}                       & Chart & Doc  & Info & Trins & Avg  \\ \midrule
LLaVA-v1.6-Vicuna-13B                       & 0.00  & 0.00 & 0.00 & 0.00  & 0.00 \\
Qwen2-VL-7B                                 & 0.00  & 0.00 & 0.00 & 0.00  & 0.00 \\
Qwen2.5-VL-7B                               & 0.00  & 0.00 & 0.00 & 0.00  & 0.00 \\
Qwen3VL-8B                                  &  8.27     & 31.19     &  9.30    & 65.16      &  28.48    \\
Qwen3VL-30B-a3b-thinking                   &  12.05    &  \underline{39.07}    & 15.49     &  \textbf{68.34}     &\underline{33.74}      \\
Qwen3.5-9B                                  &  0.73     &   20.43   & 5.08     & 44.64      &  17.72    \\
Qwen3.5-27B                                  & 2.89      &  14.77    &  5.41    &   57.04    &  20.03    \\
InternVL2-8B                                & 0.00  & 0.00 & 0.00 & 0.00  & 0.00 \\
InternVL2.5-8B                              & 0.00  & 0.00 & 0.00 & 0.00  & 0.00 \\
InternVL3-8B                              &  0.72 &17.61 & 2.02 & 21.97  &  10.58\\
InternLM-XComposer2-VL-7B                   & 0.15  & 0.20 & 0.13 & 0.57  & 0.26 \\
InternLM-XComposer2-4KHD-7B                 & 1.04  & 0.10 & 0.90 & 0.14  & 0.55 \\
Monkey-Chat                                 & 0.77  & 0.19 & 0.15 & 0.45  & 0.39 \\
MiniCPM-Llama3-V 2.5                        & 0.44  & 1.40 & 0.65 & 4.96  & 1.86 \\
Gemini-2.0-Flash                             & 0.00  & 0.00 & 0.00 & 0.00  & 0.00 \\
Gemini-2.5-Flash                             & 0.00  & 0.00 & 0.00 & 0.00  & 0.00 \\
Gemini-3-Flash                            & 11.81  & 22.35 & 14.70 & 37.03  & 21.47 \\
Gemini-3-Pro                             & \underline{18.86}  & 30.94 & \underline{20.82} & 50.42  & 30.26 \\
GPT-4o                                      & 3.90  & 1.79 & 1.60 & 13.73 & 5.26 \\
GPT-5.2                      &    3.09  &  1.87   &  2.17   &   11.98    &  4.78   \\ \midrule
\textbf{PreciseDoc-Reasoner}                                        &   \textbf{38.69}
    &  \textbf{49.81}
    &  \textbf{29.66}
    &   \underline{67.50}
   &   \textbf{46.42}   \\ \bottomrule
\end{tabular}
\end{table}

% \usepackage{booktabs}
% \usepackage{multirow}
\begin{table}[]
\centering
\scriptsize
\caption{Performance comparison of grounding and answer accuracy on BBox-DocVQA. Good Ratio denotes the ratio of generated samples containing at least one bounding box coordinate. The number of samples for SPSBB, SPMBB, MPMBB is 749, 556 and 318, respectively.}
\label{bbox_docvqa_table}
\begin{tabular}{@{}l|ccccc|cccc@{}}
\toprule
\multicolumn{1}{c|}{\multirow{2}{*}{Model}} & \multicolumn{5}{c|}{Grounding Accuracy}                                                                           & \multicolumn{4}{c}{Answer Accuracy}                                                       \\ \cmidrule(l){2-10} 
\multicolumn{1}{c|}{}                       & Good Ratio           & SPSBB                & SPMBB                & MPMBB                & ALL                   & SPSBB                & SPMBB                & MPMBB                & ALL                  \\ \midrule
Qwen2.5VL-3B                                & 51.4                 & 3.80                 & 5.60                 & 4.90                 & 4.70                  & 31.38                & 33.27                & 25.47                & 30.87                \\
Qwen2.5VL-7B                                & 73.0                 & 14.60                & 8.80                 & 8.00                 & 11.30                 & 57.01                & 53.42                & 35.53                & 51.57                \\
Qwen2.5VL-32B                               & 94.0                 & 22.20                & 20.40                & 14.30                & 20.00                 & 67.69                & 66.55                & 48.11                & 63.46                \\
Qwen2.5VL-72B                               & 99.0                 & 40.10                & 33.20                & 27.20                & 35.20                 & 71.03                & 71.58                & 57.86                & 68.64                \\
Qwen3VL-4B                                  & 95.9                & 18.90                & 17.00                & 21.30                & 18.70                 & 70.49                & 72.12                & 59.75                & 68.95                \\
Qwen3VL-8B                                  & 92.7                 & 17.60                & 9.20                 & 15.90                & 14.40                 & 70.23                & 73.20                & 51.57                & 67.59                \\
Qwen3VL-32B                                 & 94.5                 & 22.60                & 17.30                & 21.00                & 20.40                 & 81.04                & 84.35                & 55.35                & 77.14                \\
Qwen3VL-235B-a22b-thinking                                  &   \textbf{100.0}               &         52.11       &        45.93          &       42.09          &       48.03           &      88.25           &      88.13           &      84.59           &     87.49
            \\
Qwen3.5-27B                                  &   93.5               &  56.27               &   46.01               &       17.20          &    44.50              &  87.76               &    88.25             & 14.87                &    72.89             \\
InternVL3-2B                                & 80.9                & 0.00                 & 0.20                 & 0.10                 & 0.10                  & 39.12                & 33.27                & 24.21                & 34.20                \\
InternVL3-8B                                & 98.0                & 0.10                 & 0.50                 & 0.50                 & 0.30                  & 53.14                & 53.06                & 39.62                & 50.46                \\
GPT-5                                       & \underline{99.9}                 & 0.90                 & 0.10                 & 1.60                 & 1.20                  & 82.64                & 83.63                & 74.84                & 81.45                \\
GPT-5.2                                    & 99.6                & 0.88
                 & 3.31
                 & 3.39                 & 2.21                 & \textbf{90.92}                & \textbf{93.53}                & \textbf{91.51}                & \textbf{91.92}                \\
Gemini-3-Flash                                       & 99.4                 & \textbf{63.56}                 & \textbf{56.42}                 & \textbf{59.11}                 & \textbf{60.24}                  & 85.31                & 78.96                & 68.55                & 79.85                \\
Gemini-3-Pro                                       & 99.6                 & 54.70                 & 45.18                 & 37.87                 & 48.15                  & 74.36                & 46.40                & 39.62                & 57.98                \\
Doubao-Seed-1-8                            & 97.2  & 55.07 &\underline{47.91} & 37.85 & 49.25 & \underline{89.98} & \underline{90.11} & \underline{89.94} & \underline{90.02}
\\ \midrule
PreciseDoc-Reasoner                                            & \textbf{100.0} &  \underline{58.93}& 41.01 & \underline{50.61} & \underline{51.16}
 & 78.10 & 77.34 & 68.55 & 75.97 \\ \bottomrule
\end{tabular}
\end{table}

\textbf{Visual Document Grounding.}
Table \ref{doclocal4k_table} presents results on DocLocal4K that evaluates basic text grounding ability. As shown in the table, current SOTA LMMs generally exhibit poor performance with the highest score of 54.45 from Doubao-Seed-2.0-Pro. A closer examination on the figures shows generally weak localization capability on small-granularity elements such as word, phrase, and line, while those elements provides crucial evidence and higher-precision visualization. On the contrary, training with document localization data endows PreciseDoc and DocOwl-1.5 significantly stronger grounding capability across all granularities. Moreover, compared with DocOwl-1.5 that trained with a large grounding dataset DocStruct4M, our high-quality grounding data from PDF files and synthetic hand-filled documents further enhances the performance of PreciseDoc, especially on elements of line and paragraph.

The proposed PreciseDoc is capable of solving real-world grounding tasks other than the straightforward text grounding in DocLocal4K. For example, it could locate multiple critical information in the given document. %Details and demonstrations are provided in Appendix \ref{appendix:realworld}.

\textbf{Visual Grounded Reasoning.}
Table \ref{trig_table} presents results on TRIG that evaluates document understanding with word-level evidence grounding. As shown in the table, for both proprietary and open-source models, only the recent versions have acquired the ability to perform reasoning with word-level grounding. For example, Qwen series, InternVL series and Gemini begin to achieve certain results starting from the third generation, showing the growing emphasis from the community that has been placed on this task. From the table, our proposed PreciseDoc-Reasoner shows significant superiority over other baselines on this dataset, especially on the chart sub-set where it achieves twice the performance of the second-best model.

Table \ref{bbox_docvqa_table} presents results on BBox-DocVQA that evaluates document understanding with element-level evidence grounding. As shown in the table, proprietary and open-source models generally achieve better grounding scores compared to TRIG, attributed to the larger size of the document elements and the rigid layout of academic papers. From the table, our proposed model exhibits superior grounding accuracy than all benchmarked open-source LMMs, even the 235B Qwen3VL model. Compared with private models, our model is next only to Gemini-3-Flash. In terms of answer accuracy, PreciseDoc-Reasoner outperforms all listed open-source models of similar parameter size, and a subset of strong closed-source models, such as Gemini-3-Pro. There are a number of fail cases that the PreciseDoc-Reasoner correctly predicts the evidence bboxes, but encounters understanding issues, mostly seen in tables and images. We expect that extending our training pipeline to larger parameter models has the potential to achieve even better results.

\subsection{Traditional Document Understanding Benchmarks}

\setlength{\intextsep}{0pt}
\begin{wraptable}{r}{0.45\textwidth}
   \centering
   \scriptsize
   \caption{Performance comparison on traditional document understanding benchmarks.}
    \label{traditional_table}
   \begin{tabular}{@{}lccc@{}}
\toprule
Model & DocVQA & ChartQA & TextVQA \\ \midrule
Qwen2-VL-7B      &  94.5      &   \textbf{83.0}      &   \textbf{84.3}      \\
Qwen3VL-8B      &   \textbf{95.3}     &  81.2       &    77.0     \\
Qwen3.5-9B      &  \underline{95.2}      &    70.7     &  77.0       \\
DocOwl-1.5-Chat      &   82.2     &   70.2      &    68.6     \\
DocOwl-2      &   80.7     &    70.0     &    66.7     \\ \midrule
PreciseDoc-Reasoner  &  90.4      &     \underline{81.2}    &   \underline{78.2}      \\ \bottomrule
\end{tabular}
  \end{wraptable}

To assess the overall document understanding capability of the proposed PreciseDoc-Reasoner, we test it on three traditional document understanding benchmarks: DocVQA for document comprehension, ChartQA for chart understanding and TextVQA for natural images containing texts. For metrics, we use ANLS for DocVQA and text-matching accuracy for ChartQA and TextVQA. From Table \ref{traditional_table}, we can observe that PreciseDoc-Reasoner achieves competitive performance across all 3 benchmarks. In ChartQA and TextVQA, it achieves better performance than Qwen3VL-8B and Qwen3.5-9B. It is worth noting that PreciseDoc-Reasoner is specially optimized for visual grounded reasoning and does not involve intensive training on massive VQA datasets like Qwen and DocOwl series models.

\subsection{Ablation Studies}

\begin{wraptable}{r}{0.55\textwidth}
   \centering
   \scriptsize
   \caption{Ablation study of the components in training PreciseDoc-Reasoner, including the cold start and the length penalty in RL.}
   \label{ablation_reasoning_table}
   \begin{tabular}{@{}lccc@{}}
\toprule
                    & \multicolumn{1}{c|}{TRIG}    & \multicolumn{2}{c}{BBox-DocVQA} \\ \cmidrule(l){2-4} 
                      & \multicolumn{1}{c|}{Avg-IoU} & Avg-IoU          & Acc          \\ \midrule
\ding{172}\hspace{0.5em}GLM-4.6V-9B                                   &              20.81                &  28.68                &   6.33           \\
\ding{173}\hspace{0.5em}Cold-Start                              &                42.94              &        48.26          &    69.01          \\
\ding{174}\hspace{0.5em}PreciseDoc-Reasoner                                   &     \textbf{46.42}                         &   \textbf{51.16}              &      \textbf{75.97}        \\
%\ding{175} \hspace{2em}$w/o$ \hspace{0.5em}Visual Grounding RL                                             &                              &                  &              \\
%\ding{176} \hspace{2em}$w/o$ \hspace{0.5em}Grounding Reward $R_{IoU}$                                        &                  &              \\
\ding{175} \hspace{2em}$w/o$ \hspace{0.5em}Length Penalty in $R_{IoU}$                                  &     10.50                         &   72.03               &    34.57          \\ \bottomrule
\end{tabular}
  \end{wraptable}

%\textbf{Our curated grounding data by the proposed document pipeline further enhances the model's basic localization ability,} as compared with \ding{174} and \ding{175}. We evaluate the effectiveness of the training data from the two proposed document generation pipelines on DocLocal4K by comparing the model which includes the data during RL and one excluding the data. From the table, with our curated data, the performance on element localization further improves than only existing training dataset.

%\textbf{Enhancing pure grounding ability benefits visual grounded reasoning,} as compared with \ding{174} and \ding{175}. We evaluate the necessity of the training for document grounding prior to visual grounded reasoning tasks by comparing the model which includes the grounding RL with the one directly perform reasoning RL from the cold-start model.

%\textbf{Visual grounded reasoning is more effective than traditional one,} as compared with \ding{174} and \ding{176}. We evaluate the effectiveness of the joint RL training of grounding and reasoning by comparing the model whose reward contains a grounding score and one determined only by answer accuracy.
\textbf{Cold-start initialization is beneficial for visual grounded reasoning,} as compared with \ding{172} and \ding{173}. By training the model on VQA data where reasoning is inserted with grounding, the model could achieve significant performance increase, which facilitates subsequent RL training.

\textbf{The length penalty is crucial for preventing redundant bounding boxes,} as compared with \ding{174} and \ding{177}. During RL training, with the length penalty removed, the model tends to generate excessively similar bboxes to match all ground-truths. From Table \ref{ablation_reasoning_table}, TRIG evaluates both precision and recall where BBox-DocVQA only evaluates recall, explaining the significant score difference. Furthermore, the enumerating behavior by removing the length penalty leads the model to produce extremely lengthy reasoning text in the response, sometimes without a final answer.

\section{Conclusion}
This paper presents PreciseDoc, an LMM designed with advanced document localization and can be further optimized for visual grounded reasoning. We first introduce two document construction pipelines for grounding data generation that produce high-quality PDFs with accurate spatial metadata.
%Through the challenging grounding data, the proposed PreciseDoc acquires its basic localization capability. 
Furthermore, we introduce a training paradigm for document understanding that includes cold-start initialization and reinforcement learning that jointly supervises grounding and reasoning. By employing an IoU reward with the Hungarian algorithm and a length penalty, this approach achieves explainability during its reasoning and improved answer accuracy at the same time. Overall, our PreciseDoc demonstrates superior performance in document localization. The reasoning model, PreciseDoc-Reasoner, achieves improved performance in evidence grounding and reasoning accuracy across various benchmarks.

\clearpage
\bibliographystyle{plainnat}
\bibliography{main}

%%%%%%%%%%%%%%%%%%%%%%%%%%%%%%%%%%%%%%%%%%%%%%%%%%%%%%%%%%%%
\clearpage
\appendix
\section{Training Details}
\label{appendix:train}
The base model we use is GLM-4.6V-9B \cite{hong2025glm}. The whole training process consists of a cold-start SFT, a RL training for document grounding (PreciseDoc) and a further RL for visual grounded reasoning (PreciseDoc-Reasoner). In the cold-start SFT, we include both information extraction data from Section \ref{method_grounding} and reasoning data from Section \ref{method_cold_start}. We convert the information extraction data into the thinking format ,$i.e.,$ contains special tokens <think and </think>, using the same method as Section \ref{method_cold_start}. After the cold start, the model is familiar with the grounding tasks and reasoning tasks, facilitating the following two RL training. In summary, the SFT data consists of 14K DOGR data, 13K BBoxDocVQA data, 12K information extraction data and 13K DocStruct4M \cite{hu2024mplug} data. The data of RL for PreciseDoc consists of 30K data of the two grounding tasks.

%
%\section{Real-world grounding capabilities}
%\label{appendix:realworld}

%\begin{figure}[htbp]
%    \centering
%    \begin{minipage}{0.49\textwidth}
%        \centering
%        \includegraphics[width=\textwidth]{demo1.png}
%    \end{minipage}
%    \hfill
%    \begin{minipage}{0.49\textwidth}
%        \centering
%        \includegraphics[width=\textwidth]{demo2.png}
%    \end{minipage}
%    \caption{The task is to locate all elements related to "Job Experience" in the given CVs. Our PreciseDoc is capable of processing both English and Chinese documents.}
%    \label{fig:two_images}
%\end{figure}

\section{Limitations}
\label{appendix:limit}
The paper presents effective training data generation and training pipeline for the task of visual grounding and visual grounded reasoning, resulting in two models, PreciseDoc and PreciseDoc-Reasoner. The reasoning model, after the second RL, demonstrates superior visual grounded reasoning capability, and inevitably experiences slight performance degradation on previous task (document grounding). Our primary goal is to demonstrate the effectiveness of the training data and pipeline for the target tasks, and multi-task RL training is outside the scope of this work.

\end{document}